\documentclass[runningheads]{llncs}
\usepackage{graphicx}
\usepackage{comment}
\usepackage{amsmath,amssymb} %
\usepackage{color}
\usepackage{algorithm}
\usepackage{algpseudocode}
\usepackage{tabulary,multirow,overpic,xcolor}
\usepackage{makecell}

\newcolumntype{C}[1]{>{\centering}m{#1}}

\begin{document}
\pagestyle{headings}
	\mainmatter
	\def\ECCVSubNumber{}  %
	
	\title{Weakly Supervised Temporal Action Localization with Segment-Level Labels} %

	\titlerunning{}
	\author{ Xinpeng~Ding$^{1}$,
        Nannan~Wang$^{1}$,
        Xinbo~Gao$^{1}$,
        Jie~Li$^{1}$,
        Xiaoyu~Wang$^{2}$ and
        Tongliang~Liu$^{3}$}
	\authorrunning{Ding et al.}
	\institute{$^1$ Xidian University, China \\
$^2$IntelliFusion, China\\
$^3$The University of Sydney,Australia}
	\maketitle
	
\begin{abstract}
Temporal action localization presents a trade-off between test performance and annotation-time cost. Fully supervised methods achieve good performance with time-consuming boundary annotations. Weakly supervised methods with cheaper video-level category label annotations result in worse performance. In this paper, we introduce a new segment-level supervision setting: segments are labeled when annotators observe actions happening here. We incorporate this segment-level supervision along with a novel localization module in the training. Specifically, we devise a partial segment loss regarded as a loss sampling to learn integral action parts from labeled segments. Since the labeled segments are only parts of actions, the model tends to overfit along with the training process. To tackle this problem, we first obtain a similarity matrix from discriminative features guided by a sphere loss. Then, a propagation loss is devised based on the matrix to act as a regularization term, allowing implicit unlabeled segments propagation during training. Experiments validate that our method can outperform the video-level supervision methods with almost same the annotation time.
\end{abstract}
\section{Introduction}
\label{introduction}
\label{sec:introduction}
\begin{figure}[t]
\centering
\includegraphics[width=0.8\columnwidth,height=0.3\textheight]{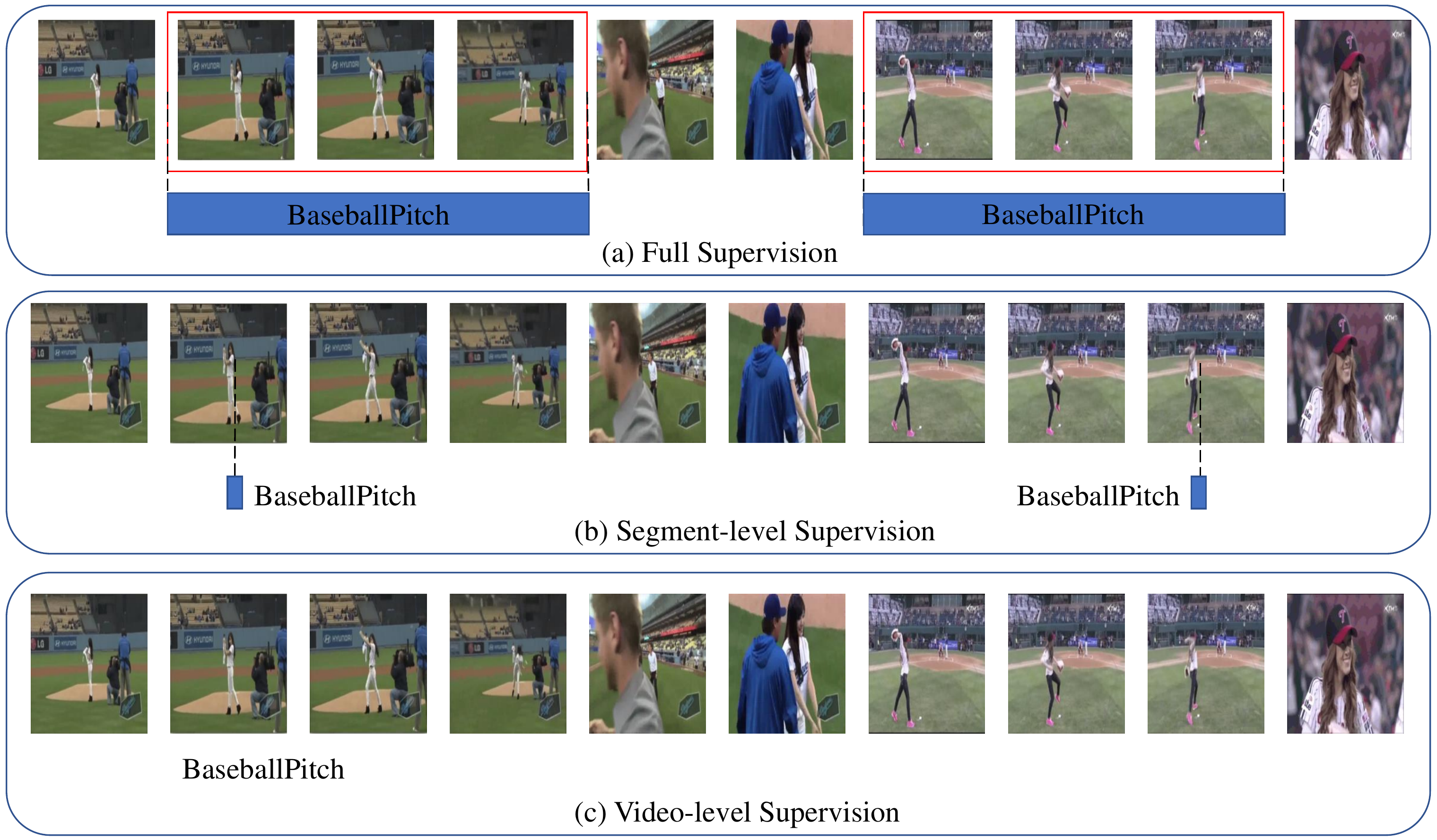} 
\caption{A video annotated with (a) full supervision, (b) segment-level supervision and (c) video-level supervision.}
\label{fig:annotate_compare}
\end{figure}

There are many works \cite{wang2017untrimmednets,chao2018rethinking,lin2018bsn,alwassel2018action,zeng2019graph,lin2019bmn} in recent years to tackle temporal action localization which aims to localize and classify actions in untrimmed videos. These methods are introduced with {\bf full supervision} setting: annotations of temporal boundaries (start time and end time) and action category labels are provided in the training procedure as shown in Fig.~\ref{fig:annotate_compare} (a). Although great improvement has been gained under this setting, obtaining such annotations is very time-consuming in long untrimmed videos \cite{article}. 

To alleviate the requirement for temporal boundary annotations, weakly supervised methods \cite{paul2018w,nguyen2018weakly,zhong2018step,long2019gaussian,narayan20193c,lee2020background} have been developed. The most common setting is {\bf video-level supervision}: only category labels are provided for each video in training time as shown in Fig.~\ref{fig:annotate_compare} (c). In these methods, researchers aim to learn class activation sequences (CAS) for action localization using excitation back-propagation with video-level category label supervision guided by a classification loss, which are simple yet efficient for weakly supervised action localization. Along with the learning procedure, CAS will shrink to the discriminative parts of action due to the discriminative parts are capable of minimizing the action classification loss. Therefore, they are usually observed to active discriminative action parts instead of full action extent. Existing approaches \cite{singh2017hide,zhong2018step} have explored erasing salient parts to expand temporal class activation maps and pursue full action extent. Nevertheless, these methods may result in decreased action classification accuracy and incomplete semantic information of actions, due to the lack of some action parts.  

In this paper, we first divide an video into non-overlap segments, each of which contains 16 frames. Then, we propose a new {\bf segment-level supervision} setting: one or two segments and their corresponding action category labels are provided in training time as shown in Fig.~\ref{fig:annotate_compare} (b). In this setting, annotators browse the video for an action instance and simultaneously label one or two interval seconds that belong to the action instance. The segments contain the labeled seconds are regard as the ground-truth labeled segments. Compared with boundary annotations, segment annotations do not require time consumption on finding precise start and end time which is sometimes accurate to 0.1 second. Furthermore, segment-level supervision can provide extra localization information compared with video-level supervision.

To make full use of this segment-level information, we propose a localization module which consists of three loss terms: a partial segment loss, a sphere loss and a propagation loss. Compared with video-level supervision methods, the partial segment loss uses the labeled segments to {\bf learn more parts of action instances instead of just focusing on discriminative parts}. Since the labeled segments are only a part of an action instance rather than the full extent, the model will overfit along with the training process. To address the problem, we first define the segments that have high feature similarity with labeled segments as implicit segments, which is motivated by the intuition that the features of the segments belonging to the same action instance are similar. To measure the similarity between pairs of segments, we obtain a similarity matrix generated from the discriminative features. Guided by the sphere loss, the discriminative features have {\bf smaller maximal intra-class distance than minimal inter-class distance}. Then, based on the obtained similarity matrix, the propagation loss is introduced to {\bf act as a regularization term which propagates labeled segments to implicit ones}. The main contributions of this paper are as follows:
\begin{itemize}
\item A new segment-level supervision setting is proposed for weakly supervised temporal action localization, costing almost the same annotation time as the video-level supervision. 
\item A novel localization module guided by a sphere loss, a partial segment loss and a propagation loss is proposed to exploit both labeled and implicit segment to keep from focusing only on the discriminative parts.
\item Experimental results demonstrate that the proposed method outperforms the state-of-the-art weakly supervised temporal action localization methods with video-level supervision setting.
\end{itemize}
\section{Related Work}
\label{related work}

\noindent{{\bf Temporal Action Localization.}}
Temporal action localization in full supervision has gained significant developments in recent years \cite{lin2017single,long2019gaussian,shou2016temporal,chao2018rethinking,lin2018bsn,lin2019bmn}. However, obtaining precise temporal boundaries (start and end time) is very time-consuming in long untrimmed videos.
To reduce the time-consumption of boundary annotations, weakly supervised temporal action localization in video-level category label supervision has attracted growing attentions. Given only category labels, most of methods \cite{wang2017untrimmednets,nguyen2018weakly,singh2017hide,zhong2018step,paul2018w,liu2019completeness} tend to generate class activation sequences (CAS) from a classification loss. However, CAS guided by the classification loss is observed to shrink to salient parts instead of the full action extent. The reason behind the phenomenon lies in that the networks tend to learn the most compact features to distinguish different categories when optimizing the classification loss and ignore less discriminative ones \cite{yuan2019marginalized}. Several researchers have attempted to pursue the integral action extent. Hide-and-Seek \cite{singh2017hide} randomly hide parts of videos during training to observe whole parts. Zhong et al. \cite{zhong2018step} trains multiple classifiers to erase regions step by step. However, these methods may lead to the lack of the discriminative parts, which would decrease the classification accuracy.
\begin{figure}[t]
\centering
\includegraphics[width=0.95\columnwidth,height=0.3\textheight]{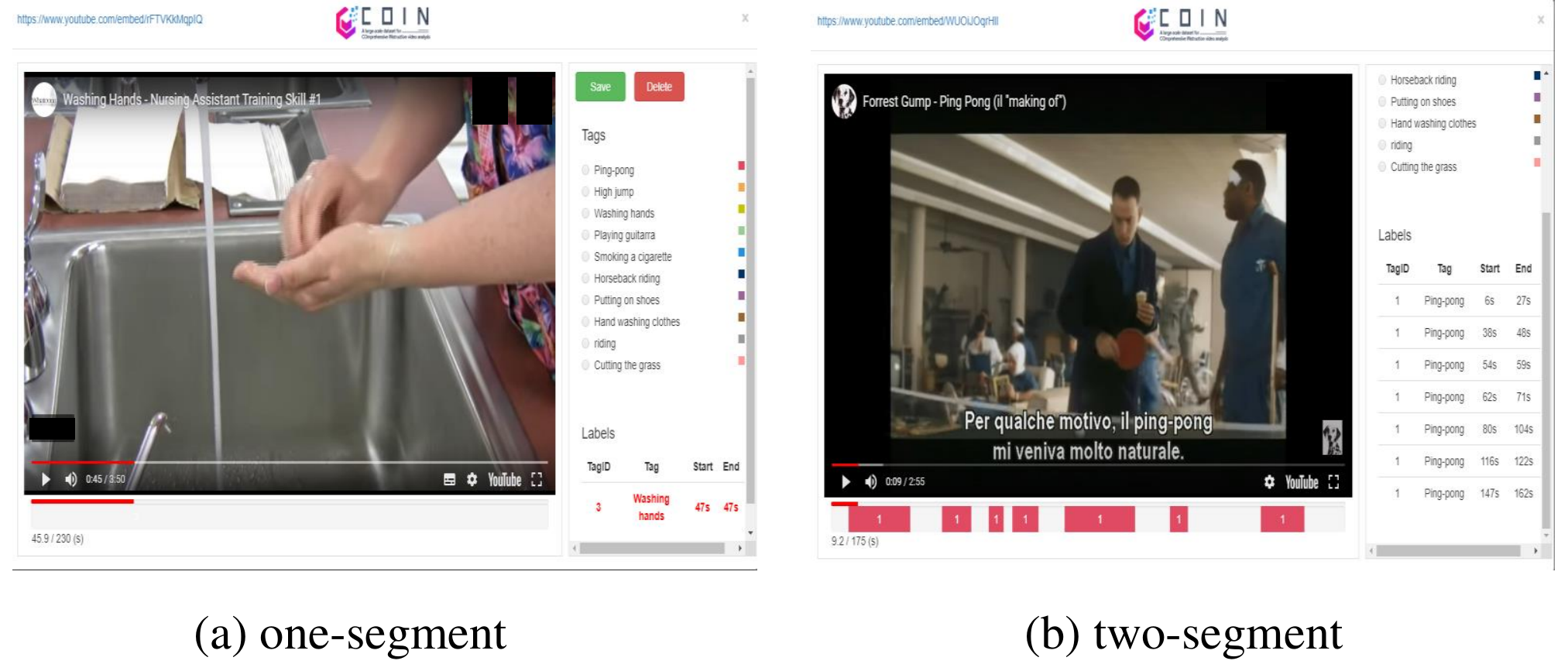}
\caption{Annotation examples of (a) one-segment and (b) two-segment with the COIN annotation tool.}
\label{fig:annotation_example}
\end{figure}

\noindent{\bf Regularization for Neural Networks.}
Regularization is a set of techniques that can prevent overfitting in neural networks and has been widely used to improve the performance, e.g. norm regularization \cite{goodfellow2016deep}, dropout \cite{srivastava2014dropout}. Motivated from the semi-supervised learning \cite{weston2012deep}, our proposed propagation loss differs from these regularization in parameters. Similar to our segment-level supervision, semi-supervised learning is the setting that a small amount of data is labeled while a large amount of ones is unlabeled during training. Weston et al. \cite{weston2012deep} add a semi-supervised loss (regularizer) to the supervised loss on the entire network’s output for unlabeled data. Such regularization is well coupled with our segment-supervised loss to improve temporal action localization performance.
\begin{figure}[t]
\centering
\includegraphics[width=0.95\columnwidth,height=0.3\textheight]{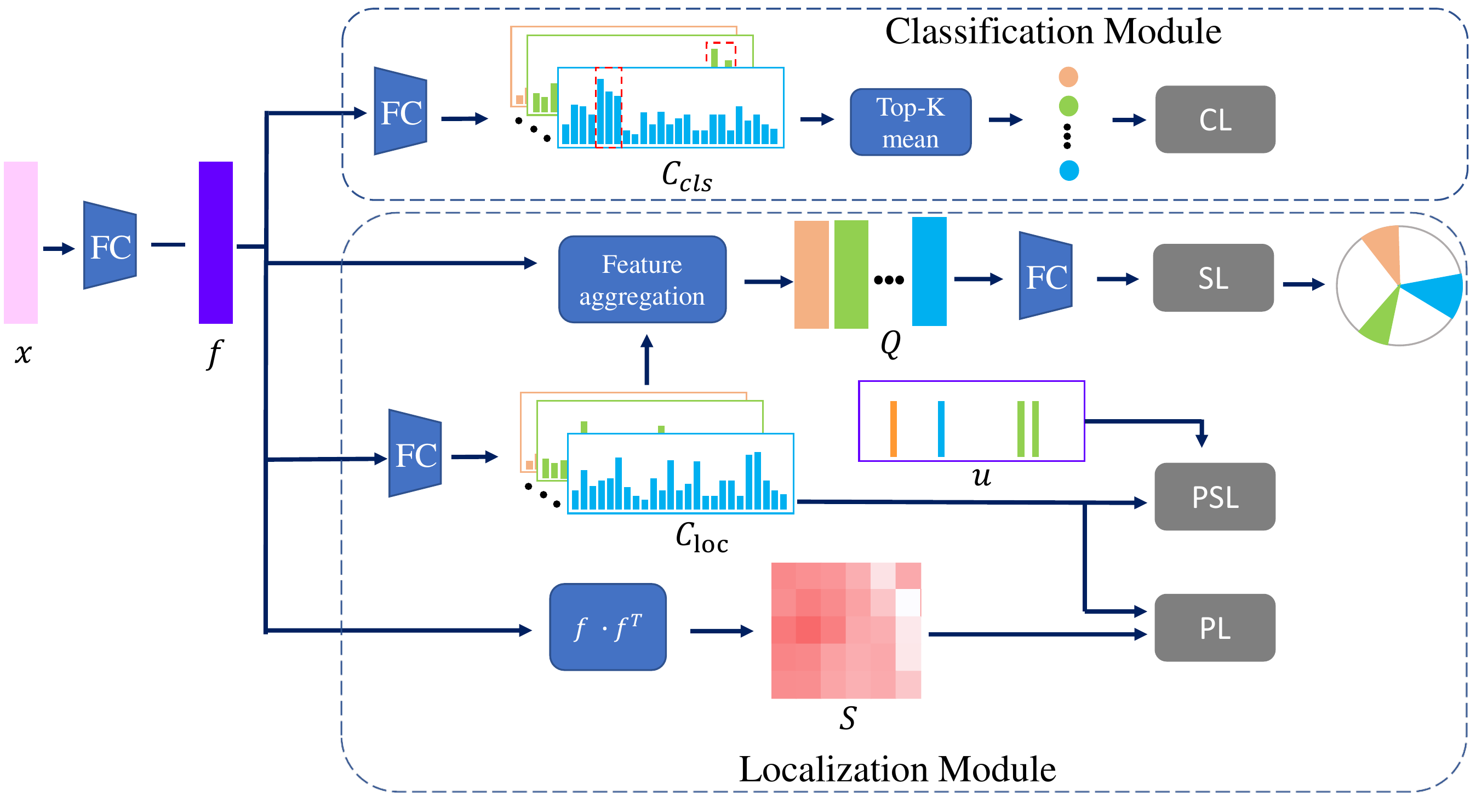}
\caption{Architecture of Our Approach. There are two main modules: a classification module and a localization module. The classification module is trained by a classification loss (CL) and the localization module is guided by a partial segment loss (PSL), a sphere loss (SL) and a propagation loss (PL).}
\label{fig:architecture}
\end{figure}
\section{Annotation of Segment-Level Supervision}
\label{sec:segmentlabel}
For preventing re-annotations, we generate ground-truth segment labels by random sampling available temporal action boundary annotations from ActivityNet and THUMOS14. However, for a new dataset, segment labels can be annotated, without demand of any action boundary annotations. We choose two kinds of segment-level supervision: {\bf one-segment} in which one segment is labeled for each action instance and {\bf two-segment} in which two interval segments are labeled for each action instance. In two-segment, since temporal annotations are one-dimensional, segments between two labeled interval segments can all be regard as ground-truth segments. We sample $50$ videos with $10$ action classes from ActivityNet and THUMOS14 for evaluating annotation time. We use the COIN annotation tool \cite{coin} to label the seconds in which an action instance happens, as shown in Fig.~\ref{fig:annotation_example}. Then the labeled ground-truth segments can be regarded as the segments which contain the labeled seconds. The experiments on annotation time of video-level, one-segment, two-segment and full supervision are conducted in Section \ref{sec:exploratory experiments}.

\section{Our approach}
\label{sec:approach}

\subsection{Problem Statement and Notation}
\label{sec:ps}
Let define untrimmed videos as $V=\left \{ v_ i\right\}_{i=1}^M$, where $M$ denotes the number of videos. We divide each video into non-overlap segments $\left \{ {g_t} \right \}_{t=1}^{l_i}$, where $l_i$ denotes number of segments. Each segment consists of $16$ frames. The extracted feature of $v_i$ is denoted as $x_i \in \mathbb{R}^{l_i \times D}$, where $D$ is the dimension. Let the action label be denoted as $Y=\left\{y_{i}\right\}_{i=1}^{M}$, where $y_i \in\{0,1\}^{N}$ is a multi-hot vector and $N$ is the number of action classes. For the video $v_i$, we denote its segment label as $u_i \in \{0,1\}^{l_i \times N}$. $u_i(t,n)=1$ when there is an action instance with category $n$ occurring in $t$-th segment and $u_i(t,n)=0$ when none action instance with category $n$ occurs in the $t$-th segments. For simplicity, we use $x$, $u$ and $l$ instead of $x_i$, $u_i$ and $l_i$ when there is no confusion. We use $Q(i,j)$ to represent the elements in the $i$-th row and $j$-th column of matrix $Q$. Naturally, $Q(i,:)$ and $Q(:,j)$ indicate the $i$-th row vector and $j$-th column vector of matrix $Q$ respectively.  
\subsection{Architecture}
\label{sec:architecture}
The architecture of our approach is shown in Fig.~\ref{fig:architecture}. The fused feature $x$ described in Section \ref{sec:ps} is fed into a fully connected (FC) layer to get the discriminative feature $f \in \mathbb{R}^{l \times D}$. Following is two main modules: a classification module to learn discriminative parts for distinguishing different action classes and a localization module to observe integral action regions. 

The output of a fully connected layer in the classification module is the class activation sequence (CAS) which is a class-specific 1D temporal map, similar to the 2D class activation map in object detection \cite{zhou2016learning}. We denote the CAS for classification as $C_{cls} \in \mathbb{R}^{l \times N}$. Conducting the top-k mean operation on $C_{cls}$, a probability mass function (PMF), denoted by $p$, is generated for a classification loss. Similar to other video-level supervision methods \cite{paul2018w,nguyen2018weakly}, the classification loss encourages the model to distinguish the different action categories.

In the localization module, we obtain a localization CAS, denoted by $C_{loc} \in \mathbb{R}^{l \times N}$, with a fully connected layer similar to the classification module. Guided by a partial segment loss, the model can pay attention to labeled segments rather than only the discriminative ones learned from the classification loss. Since the labeled segments are only part of action instances, the model are prone to overfit as the training proceeds. To solve this drawback, we define the segments having high similarity with labeled segments as the implicit segments. In order to measure the similarity of pairs of segments, a sphere loss is first adopted to ensure $f$ has smaller maximal intra-class distance than minimal inter-class distance. Then, we measure the similarity between pairs of segments by the similarity matrix $S=f \cdot f^T$, where $\cdot$ and $f^T$ indicates the matrix multiplication and the transpose of $f$ respectively. Finally, we add a propagation loss to propagate $C_{loc}$ over partially labeled segments to the entire action instances including unlabeled implicit segments. The objective of our framework is formulated as follows:
\begin{equation}
    \mathcal{L}=\mathcal{L}_{cls}+\alpha \mathcal{L}_{segment}+\beta \mathcal{L}_{sphere} + \gamma \mathcal{L}_{prop},
    \label{e:loss}
\end{equation}
where $\mathcal{L}_{cls}$, $\mathcal{L}_{segment}$, $\mathcal{L}_{sphere}$ and $\mathcal{L}_{prop}$ indicate the classification loss, the partial segment loss, the sphere loss and the propagation loss respectively. $\alpha$, $\beta$ and $\gamma$ are trade-off hyperparameters.
\begin{figure}[t]
\centering
\includegraphics[width=0.95\columnwidth,height=0.22\textheight]{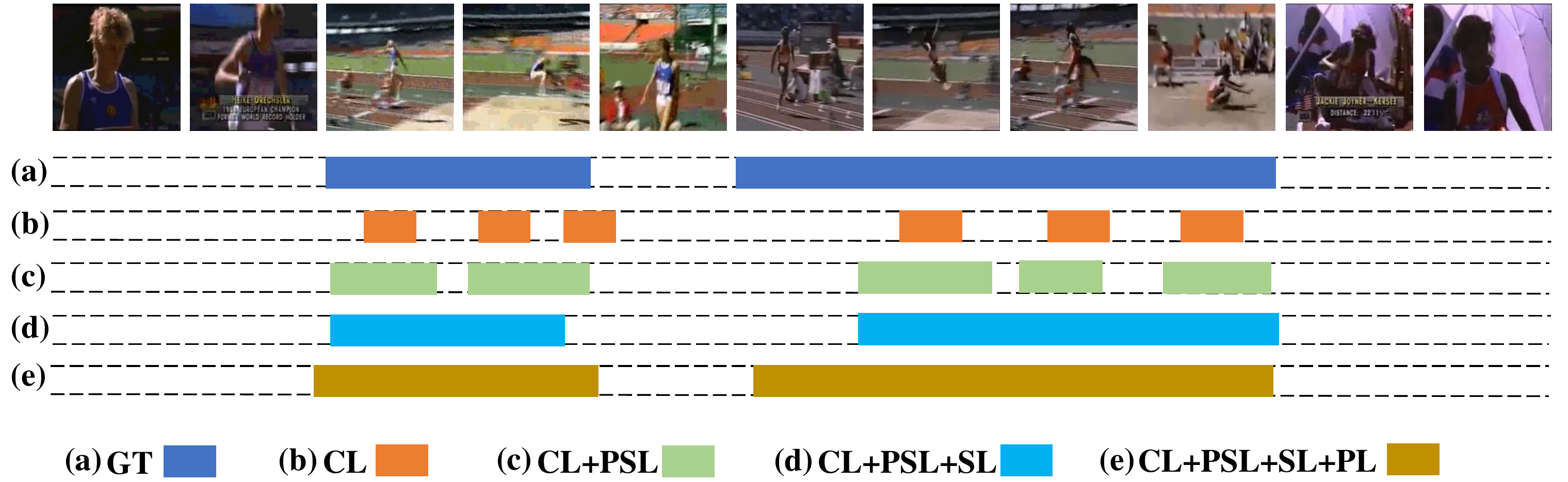}
\caption{Predicted action proposals for a video clip containing `LongJump' category from THUMOS14. (a) `GT' indicates the ground-truth segments belonging to action instances. (b) The model trained with the classification loss (CL) only predict the discriminative segments. (c) Trained with the classification loss (CL) and partial segment loss (PSL), the model can observe more segments belonging to the action instances. (d) After adding the sphere loss (SL), the segments belonging to the same action instance are combined. (e) Guided by the propagation loss (PL), more implicit segments belonging to action instances are predicted.}
\label{fig:cas_compared}
\end{figure}
\subsection{Classification Loss}
\label{sec:cl}

Due to the variation temporal duration, we use the top-k mean to generate a single class score aggregated from $C_{cls}$ described in Section \ref{sec:architecture}, similar to \cite{paul2018w}. The class score for the $n$-th category, denoted by $s^n$, is defined as follows:
\begin{equation}
   s^{n}=\frac{1}{k} \max _{\mathcal{M} \subset C_{cls}(:,n)} \sum_{m \in \mathcal{M}} m,
    \label{e:classscore}
\end{equation}
where $|\mathcal{M}|=k$, $k=\left\lfloor \frac{T}{r} \right\rfloor$ and $r$ is a hyperparameter to control the ratio of selected segments in a video. Then, a probability mass function (PMF), $p^n$, is computed by employing softmax:
\begin{equation}
p^{n}=\frac{\exp \left(s^{n}\right)}{\sum_{n=1}^{N} \exp \left(s^{n}\right)}.
    \label{e:pmf}
\end{equation}
Finally, the classification loss (CL) is defined as follows:
\begin{equation}
  \mathcal{L}_{cls}=\frac{1}{N}  \sum_{n=1}^{N}-y^{n} \log \left(p^{n}\right),
    \label{e:classloss}
\end{equation}
where $y^n$ is the ground-truth label for $n$-th class which described in Section \ref{sec:ps}. 

Along with the training process, the CAS guided by the classification loss will shrink to only the discriminative parts rather than the whole action instances. Specific activated action parts are capable of minimizing the action classification loss, but difficult to optimize action localization. {\it The only goal of optimizing CL is to capture the relevant action parts between $f$ and $y$ to distinguish action categories. Along with training, the relevant parts become more and more discriminative while the irrelevant parts with no contribution to the prediction of $y$ are suppressed.} As illustrated in Fig.~\ref{fig:cas_compared} (b), for `LongJump' category, only the segments where the athlete jumps from the bunker are predicted. This is because these parts can be informative and enough to distinguish `LongJump' from other different action categories, such as `HighJump'.
\subsection{Partial Segment Loss}
\label{sec:psl}
In order to tackle the problem described in Section \ref{sec:cl}, we introduce a partial segment loss in segment-level supervision. An intuitive choice is $\ell_2$ loss, denoted by $\ell_2=\sum_{n=1}^{N}\sum_{t=1}^{l} \left \| C_{loc}(t,n)-u(t,n) \right\|^2$. 
The model guided by $l_2$ will urge $C_{loc}$ to fit $u$.
However, the ground-truth labels $u$ are only the partial action instances rather than the entire ones. Therefore, We introduce a partial segment loss which only considers the cross entropy loss for labeled segments $u$ and effectively ignores other parts. We first conduct softmax on $C_{loc}$ to obtain the normalized CAS, defined as: 
\begin{equation}
   a(t,n)=\frac{\exp \left(C_{loc}(t,n)\right)}{\sum_{i=1}^{l} \exp \left(C_{loc}(i,n)\right)}.
   \label{e:normalcas}
\end{equation}
Then, the partial segment loss can be defined as follows:
\begin{equation}
   \mathcal{L}_{segment}= \sum_{(t,n) \in \Omega}-u(t,n) \cdot \log \left( a(t,n)\right),
    \label{e:psl}
\end{equation}
where $\Omega=\left \{(t,n) |u(t,n)=1 \right \}$. This partial segment loss can be seen as a sampling of the loss which is consistent with the annotation of the segment-level supervision. In segment-level supervision, the process of labeling segments can be regarded as a sampling of action instances. Guided by the partial segment loss, the model can observe more essential parts which is shown in Fig.~\ref{fig:cas_compared} (c).
\subsection{Sphere Loss}
\label{sec:sl}
As described in Section \ref{sec:architecture}, we denote the similarity matrix between two segments as $S=f \cdot f^T$. To ensure features with the same category have higher similarity than those with different categories, $f$ should have the property that maximal intra-class distance is smaller than minimal inter-class distance. 
The A-Softmax loss introduced in \cite{liu2017sphereface} learns the features by constructing a discriminative angular distance metric, making the decision boundary more stringent and separated. However, A-Softmax loss is used for face recognition, which is trained on examples containing single-label instances with no backgrounds. 

In our task, we integrate the sphere loss adapted from A-Softmax loss \cite{liu2017sphereface} into our network for multi-label action instances. Since an untrimmed video contains many background clips, a feature aggregation is needed to obtain a class-specific feature without background regions. Specifically, let $\tau^{n}=median\left(a(:,n)\right)$ and we first compute the high attention along the temporal axis for class $n$ as follows:
\begin{equation}
 A\left(t, n\right)=\left\{\begin{array}{ll}
{a\left(t,n\right)} & {,\text { if } a\left(t,n\right) \geq \tau^n} \\
{0} & {,\text { if } a\left(t,n\right)< \tau^n}
\end{array}\right.,
    \label{e:attention}
\end{equation}
where $a\left(t,n\right)$ is the normalized CAS in Equation \ref{e:normalcas}. We refer to $A$ as attention, as
it attends to the portions of the video where an action of a certain category occurs. For example, if $A(t,n)$ equals $a(t,n)$ instead of $0$, the $t$-th segments of the video may contain action instances of category $n$.
Then as in \cite{paul2018w}, we obtain the high attention region aggregated class-wise feature vectors for category $n$ as follows:
\begin{equation}
 F^n=\left(f \right)^T \cdot A(:,n),
    \label{e:classwisef}
\end{equation}
where $F^n \in \mathbb{R}^{D}$. 
Following \cite{liu2017sphereface}, we define the fully connect layer as $W$ and $\theta_{i,j}$ is the angle between $W(j)$ and $F^n(i)$. Then, the A-Softmox loss for category $n$ is formulated as:
\begin{equation}
\mathcal{L}^n_{ang}=\frac{1}{D} \sum_{i=1}^D-\log \left(\frac{e^{\left\|F^n(i)\right\| \psi\left(\theta_{n, i}\right)}}{e^{\left\|F^n(i)\right\| \psi\left(\theta_{n, i}\right)}+\sum_{j \neq n} e^{\left\|F^n(i)\right\| \cos \left(\theta_{j, i}\right)}}\right),
    \label{e:asoftmaxloss}
\end{equation}
where $\psi\left(\theta_{n, i}\right)=(-1)^{k} \cos \left(m \theta_{n, i}\right)-2$, $\theta_{n, i} \in\left[\frac{k \pi}{m}, \frac{(k+1) \pi}{m}\right]$ and $k \in[0, m-1]$. $m \geq 1$ is an integer that controls the size of angular margin. More detail explanation and provement can be found in \cite{liu2017sphereface}. Then, the sphere loss for multi-label action categories can be formulated as:
\begin{equation}
\mathcal{L}_{sphere}=\sum_{n=1}^N \mathcal{L}^n_{ang}.
    \label{e:sphereloss}
\end{equation}
The predicted proposals of added sphere loss is shown in Fig.~\ref{fig:cas_compared} (d). 
\subsection{Propagation Loss}
\label{sec:pl}
In segment-level supervision, only a part of action instances is labeled, compared with entire regions in full supervision methods. This setting is similar to the semi-supervised learning which combines a small amount of labeled data with a large amount of unlabeled data during training. In many semi-supervised algorithms \cite{chapelle2003cluster,zhu2002learning}, a key assumption is the structure assumption: points within the same
structure (such as a cluster or a manifold) are likely to have the same label. Under this assumption, the aim is to use this structure to propagate labeled data to unlabeled data. In \cite{weston2012deep}, authors add a semi-supervised loss (regularizer) to the supervised loss on the entire network’s output:
\begin{equation}
\sum_{i=1}^{L} \ell\left(E\left(x_{i}\right), y_{i}\right)+\gamma \sum_{i, j=1}^{L+U} H\left(E\left(x_{i}\right), E\left(x_{j}\right), W_{i j}\right),
    \label{e:semiloss}
\end{equation}
where $L$ and $U$ indicate the number of the labeled and unlabeled examples respectively. $E$ indicates the encoding function and $W_{ij}$ specifies the similarity or dissimilarity between examples $x_i$ and $x_j$. $\ell$ is the loss for labeled examples and $H$ is the loss between pairs of examples. $\gamma$ is the trade-off hyperparameter. In our approach, we rewrite the Equation \ref{e:semiloss} as follows:
\begin{equation}
     \underbrace{ \sum_{(t,n) \in \Omega}-u(t,n) \cdot \log \left(a(t,n)\right)}_{\text {Partial segment Loss }}+\gamma \underbrace{\sum_{n=1}^N \mathcal{L}_{prop}\left( C_{loc}^{n}, S\right)}_{\text {Regularizer}},
    \label{e:rewriteloss}
\end{equation}
where the propagation loss is defined as follows:
\begin{equation}
\mathcal{L}_{prop}\left(C_{loc}^{n}, S\right)= \sum_{i,j=1}^{l}S \left\| C_{loc}^{n}(i)- C_{loc}^{n}(j)\right\|^2,
    \label{e:propagationloss}
\end{equation}
where $S$ is the similarity matrix described in Section \ref{sec:architecture}. With the propagation loss, the model can propagate the labeled segments to implicit segments by measuring their similarity, as shown in Fig.~\ref{fig:cas_compared} (e).
\subsection{Classification and Localization}
\label{sec:cal}
We first get the final CAS, denoted by $C_a= \frac{C_{cls}+ C_{loc}}{2}$. Then, $s_a$ and $p_a$ are computed by Equation \ref{e:classscore} and \ref{e:pmf}. As in \cite{paul2018w}, we use the computed PMF $p_a$ with a threshold to classify the video to contain one or more action categories. For localization, we discard the categories of which $s_a$ are below a certain threshold (0 in our experiments). Thereafter, for each of the remaining categories, we apply a threshold to $C_a$ along the temporal axis to obtain the action proposals.
\begin{table}[t]
\begin{center}
\caption{Results (mAP) with different loss terms on THUMOS14 at IoU=0.5. `one-segment' and `two-segment' indicate labeling one segment and two segments for each action instance respectively.}\smallskip
\label{table:ablation}
\begin{tabular}{l|c c c c c c c c}
\hline
CL (baseline)& \quad\checkmark\quad & \quad\checkmark\quad & \quad\checkmark\quad &\quad\checkmark\quad & \quad\checkmark\quad & \quad\checkmark\quad & \quad\checkmark\quad & \quad\checkmark\quad\\
PSL& & \quad\checkmark\quad & & & \quad\checkmark\quad & \quad\checkmark\quad & & \quad\checkmark\quad\\
SL& & & \quad\checkmark\quad & & \quad\checkmark\quad & & \quad\checkmark\quad & \quad\checkmark\quad\\
PL& &  & & \quad\checkmark\quad & & \quad\checkmark\quad & \quad\checkmark\quad & \quad\checkmark\quad\\
\hline
one-segment& \quad19.4\quad & \quad27.0\quad & \quad24.4\quad & \quad19.7 \quad& \quad28.6 \quad& \quad27.2 \quad& \quad26.1 \quad& \quad {\bf 29.3}\quad \\
two-segment& \quad19.4\quad & \quad28.5\quad & \quad24.4\quad & \quad19.7 \quad& \quad29.9 \quad& \quad29.1 \quad& \quad26.1 \quad& \quad{\bf 31.6}\quad \\
\hline
\end{tabular}
\end{center}
\end{table}
\section{Experiments}
\label{sec:expriments}

\subsection{Experimental Setup}

{\bf Datasets.} We evaluate our method on two popular action localization benchmark datasets: THUMOS14 \cite{idrees2017thumos} and ActivityNet \cite{caba2015activitynet}. 
The THUMOS14 dataset has temporal annotations for a subset of videos in the validation and test sets for 20 classes. ActivityNet1.2 has 4,819 videos for training, 2,383 videos for validation, and 2,480 videos for testing whose labels are withheld. ActivityNet1.3 contains 19,994 videos with 200 action classes. 

\begin{table}[t]
\centering
\caption{Action localization performance comparison (mAP) of our method with the state-of-the-art methods on the THUMOS14 dataset. The mAP values at different IoU thresholds and the average mAP (0.1:0.1:0.5) are presented. UNT and I3D are abbreviations for UntrimmedNet features and I3D features respectively. Our two-segment model with full loss terms achieves the best performance at most IoUs with both UntrimmedNet and I3D features.}\smallskip
\label{table:thumos}
\resizebox{0.9\columnwidth}{!}{
\begin{tabular}{l|c c c c c c c c}
\hline
\hline
    \multirow{2}{*}{Methods}& \multicolumn{8}{c}{mAP @ IoU}  \\
    \cline{2-9}
  & 0.1 & 0.2 &0.3  & 0.4 & 0.5 &0.6&0.7&AVG\\
   \hline
   \hline
\multicolumn{9}{c}{Fully-supervised Methods}\\
\hline
S-CNN \cite{shou2016temporal} &47.7 &43.5 &36.3& 28.7 &19.0 &-& 5.3 &35.0\\

R-C3D \cite{xu2017r} & 54.5 &51.5 &44.8& 35.6 &28.9 &-& -& 43.1\\

SSN \cite{zhao2017temporal}& 60.3& 56.2& 50.6 &40.8& 29.1& - &- &47.4\\

Chao et al. \cite{chao2018rethinking} &59.8 &57.1 &53.2 &48.5 &42.8& 33.8& 20.8& 52.3\\
GTAN \cite{long2019gaussian} &{\bf 69.1}&{\bf 63.7} &{\bf 57.8}& {\bf 47.2}&{\bf38.8}& -& -&{\bf 55.32}\\
\hline
\hline
\multicolumn{9}{c}{Weakly-supervised Methods}\\
\hline
Hide-and-Seek \cite{singh2017hide} &36.4& 27.8& 19.5 &12.7& 6.8& - &- &20.6\\

 Zhong et al. \cite{zhong2018step}& 45.8 &39.0 &31.1& 22.5& 15.9 &-& -& 30.9\\
STPN (UNT) \cite{nguyen2018weakly} &45.3 &38.8 &31.1& 23.5 &16.2& 9.8& 5.1& 31.0\\

W-TALC (UNT) \cite{paul2018w}& 49.0 &42.8 &32.0 &26.0 &18.8 &- &6.2& 33.7\\

AutoLoc (UNT) \cite{shou2018autoloc}& - &- &35.8 &29.0 &21.2 &13.4 &5.8& -\\

Liu et al. (UNT) \cite{liu2019completeness} & 53.5 &46.8 &37.5& 29.1 &19.9 &12.3& 6.0& 37.4\\
 BaSNet (UNT) \cite{lee2020background} & 56.2 &50.3 &42.8 &34.7 &25.1 &17.1 &9.3 & 41.8\\
 \hline
Ours (UNT) & {\bf 59.1} & {\bf53.5}& {\bf45.7}& {\bf37.5} &{\bf28.4} &{\bf 20.3}& {\bf11.8} & {\bf 44.8}\\
\hline

 STPN (I3D) \cite{nguyen2018weakly}& 52.0& 44.7& 35.5& 25.8 &16.9& 9.9& 4.3& 35.0\\

 W-TALC (I3D) \cite{paul2018w} &55.2& 49.6& 40.1 &31.1& 22.8& - &7.6& 39.8\\

 Liu et al. (I3D) \cite{liu2019completeness}&57.4 &50.8& 41.2& 32.1 &23.1 &15.0& 7.0 &40.9\\
 3C-Net (I3D) \cite{narayan20193c}&59.1 &53.5 &44.2 &34.1 &26.6 &8.1&-&43.5\\
 BaSNet (I3D) \cite{lee2020background}&58.2 &52.3 &47.6 &36.0 &27.0 &18.6 &10.4&43.6\\
\hline

 Ours (I3D)  & {\bf 61.6} &{\bf55.8}& {\bf48.2}& {\bf39.7} &{\bf31.6} &{\bf22.0}& {\bf13.8} &{\bf 47.4}\\
 \hline
 \hline
\end{tabular}}
\end{table} 
\noindent{\bf Evaluation metric.} Following previous methods \cite{paul2018w,nguyen2018weakly,narayan20193c,lee2020background}, we use the standard protocol provided by two datasets  for evaluation. For action localization, the evaluation protocol is based on mean Average Precision (mAP) score at different intersection over union (IoU) values. For the multi-label action classification, we use the predicted video-level scores to compute the mAP score for evaluation.

\noindent{\bf Implementation details.} We use the corresponding repositories to extract the features
for UntrimmedNet \cite{wang2017untrimmednets} and I3D \cite{carreira2017quo}. The dimension of the confused feature $x$ is $D=2,048$. As in \cite{paul2018w,narayan20193c}, We do not finetune the feature extractors. The trade-off hyperparameters in Equation \ref{e:loss} are $\alpha=0.1$, $\beta=0.0001$ and $\gamma=0.02$ respectively. Different from previous video-level supervision methods which use a fixed number, the ratio of selected segments $r$ is set to $2Q$, where $Q$ indicates the number of labeled segments in our experiments. All of our models are implemented by PyTorch \cite{32} and trained under the environment of Python 3.6 on Ubuntu 16.04 system with a 12G NVIDIA Titan Xp GPU.

\subsection{Exploratory Experiments}
\label{sec:exploratory experiments}
In the following experiments, we take I3D \cite{carreira2017quo} as the feature extractor.

{\bf Ablation study.} We set the model guided by the classification loss (CL) alone as the baseline. The comparison of temporal action localization performance (mAP) with different loss terms on THUMOS14 at IoU=0.5 are shown in Table~\ref{table:ablation}. The baseline model gets a mAP score of $19.4\%$. The partial segment loss (PSL) can significantly improve the performance. In one-segment label, combining the classification loss and partial segment loss (CL+PSL), we can obtain a mAP score of $27.0\%$, improving $7.6\%$ over CL. In two-segment label, with more labeled segments, the performance is significantly improved, $9.1\%$ over CL. The sphere loss is also beneficial to localization due to more discriminative features generated. For instances, the integration of the classification loss and sphere loss (CL+SL) obtains a mAP score of $24.4\%$ improving $5.0\%$ over CL. With only the propagation loss (PL), the performance is hardly improved for the propagation loss requiring the similarity information. However, PL can propagate predicted regions to implicit parts based on other loss terms.
Guided by CL+PSL+PL, we obtain better performance than CL+PSL, i.e., a mAP score of $27.2\%$ for one-segment. PL can also improve $1.5\%$ over CL+SL. The action localization performance is improved to $29.3\%$ and $31.6\%$ mAP for one-segment and two-segment respectively, by CL+PSL+SL+PL.

{\bf Comparisons of the trade-off between annotation time and performance.}
To evaluate the annotation time, we define a new metric named as annotation-duration ratio which is denoted by $\phi=\frac{t_a}{l}$, here $t_a$ and $l$ indicate the annotation time and duration time of videos respectively. Using the COIN annotation tool \cite{coin} to label on the THUMOS14 dataset, we obtain $\phi_{video}=0.24$, $\phi_{segment_1}=0.30$, $\phi_{segment_2}=0.32$ and $\phi_{full}=1.17$, where $\phi_{video}$, $\phi_{segment_1}$ $\phi_{segment_2}$ and $\phi_{full}$ indicate $\phi$ of video-level, one-segment, two-segment and full supervision respectively. Similarly, for the sampled videos of ActivityNet, we obtain $\phi_{video}=0.25$, $\phi_{segment_1}=0.33$, $\phi_{segment_2}=0.38$ and $\phi_{full}=1.45$.
We present the trade-off between annotation time and performance on the THUMOS14 dataset in Fig.~\ref{fig:time_mAP}. The x-axis is the annotation time ($\phi$) which is denoted in Section \ref{sec:segmentlabel} and the y-axis is the performance (mAP) of temporal action localization. As Fig.~\ref{fig:time_mAP} indicates, our approach in segment-level supervision can significantly improve the performance compared with video-level supervision methods, at the cost of only a little more annotation time.
 
\begin{table}[t]
\begin{minipage}{0.49\linewidth}  
\centering
\caption{Action localization performance comparison (mAP) of
our method with the state-of-the-art methods on ActivityNet1.2 dataset. `AVG' means the the average mAP at IoU thresholds 0.5:0.05:0.95.}
\label{table:ant1.2}
\resizebox{0.9\textwidth}{!}{\begin{tabular}{l| c c c c}
\hline
\hline
  \multirow{2}{*}{\; Methods}& \multicolumn{4}{c}{mAP @ IoU}  \\
\cline{2-5}
  &  0.5 & 0.75 &0.95 &AVG\\
 \hline
 \hline
\multicolumn{5}{c}{Fully-supervised Methods}\\
\hline
SSN \cite{zhao2017temporal}& {\bf 41.3}& {\bf 27.0}&{\bf 6.1} &{\bf 26.6}\\
\hline
\hline
\multicolumn{5}{c}{Weakly-supervised Methods}\\
\hline
 W-TALC\cite{paul2018w}&37.0& - &- &18.0\\
Liu et al. \cite{liu2019completeness}&36.8& 22.0 &5.6 &22.4\\
3C-Net \cite{narayan20193c}&37.2 &23.7 &9.2 &21.7\\
BaSNet \cite{lee2020background}&38.5 &24.2 &5.6 &24.3\\
\hline
Ours  &{\bf41.7}& {\bf26.7} &{\bf6.3}& {\bf26.4}\\
\hline
\hline
\end{tabular}}
\end{minipage}
 \hfill
\begin{minipage}{0.49\linewidth}  
\centering
\caption{Action localization performance comparison (mAP) of
our method with the state-of-the-art methods on ActivityNet1.3 dataset. `AVG' means the the average mAP at IoU thresholds 0.5:0.05:0.95.}
\label{table:ant1.3}
\resizebox{0.9\textwidth}{!}{\begin{tabular}{l| c c c c}
\hline
\hline
 \multirow{2}{*}{\; Methods}& \multicolumn{4}{c}{mAP @ IoU}  \\
\cline{2-5}
  & 0.5 & 0.75 &0.95 &AVG\\
 \hline
 \hline
\multicolumn{5}{c}{Fully-supervised Methods}\\
\hline
GTAN \cite{long2019gaussian}& {\bf 52.6}& 34.1& {\bf 8.9} &{\bf 34.3} \\
BMN \cite{lin2019bmn}& 50.1 &{\bf 34.8} &8.3 &33.9 \\
\hline
\hline
\multicolumn{5}{c}{Weakly-supervised Methods}\\
\hline
STPN \cite{nguyen2018weakly}&29.3& 16.9 &2.6 &-\\
Liu et al. \cite{liu2019completeness}&34.0& 20.9 &5.7 &21.2\\
BaSNet \cite{lee2020background}&34.5& 22.5& 4.9& 22.2\\
\hline
Ours  &{\bf37.7}& {\bf25.6} &{\bf6.8}& {\bf24.8}\\
\hline
\hline
\end{tabular}}
\end{minipage}
\end{table}
\subsection{Comparisons with the State-of-the-art}
We conduct experiments on THUMOS14 and ActivityNet datasets to compare with several state-of-the-art techniques. 

{\bf Action localization.} Table \ref{table:thumos} reports the comparison of our method with existing approaches on the THUMOS14 dataset. We report mAP scores at different IoU thresholds (0.1:0.1:0.7). `AVG' represents the average mAP at IoU thresholds form 0.1 to 0.5. Results show that our model can perform better than previous video-level weakly supervised methods at all IoU thresholds for both UNT and I3D feature extractors. Our method can achieve better performance at most of the IoUs. Specifically, for average mAP from 0.1 to 0.5, our method significantly improves the performance from $41.8\%$ to $44.8\%$ with the UNT features. The performance of our method is further improved by using I3D features, and we achieved an average mAP of $47.4\%$, which improves $3.8\%$ over BasNet \cite{lee2020background}. Table \ref{table:ant1.2} shows the state-of-the-art comparison on the ActivityNet1.2 dataset. Following other works \cite{narayan20193c,lee2020background}, we report the mean mAP scores at different thresholds (0.5:0.05:0.95). Our approach achieves an average mAP of $ 26.4\%$ which surpasses all existing video-level weakly-supervised methods. Furthermore, our segment-level supervision also have competitive performance against the full supervision method SSN \cite{zhao2017temporal}. Table \ref{table:ant1.3} illustrates the performance comparison of our method with the start-of-the-art on ActivityNet1.3 dataset. Our method with segment-level supervision achieves an average mAP of $24.8\%$, outperforming BaSNet in video-level supervision by $2.8\%$.

{\bf Action classification.} Table \ref{table:classification} reports action classification performance comparison (mAP) of our method with the state-of-the-art methods on the THUMOS14 and ActivityNet 1.2 datasets. Since $r$ varies with the number of labeled segments in the video, it can represent more appropriate sampling information compared with the fixed number. In comparison with the existing approaches, our method achieves competitive results of $87.6\%$ and $93.2\%$ mAP on THUMOS14 and ActivityNet1.3 respectively. 

\begin{table}[t]
\makeatletter\def\@captype{table}\makeatother
\begin{minipage}{0.49\linewidth}
\centering
\caption{Action classification performance comparison of
our method with the state-of-the-art methods on the THUMOS14 and
ActivityNet 1.2 datasets.}
\label{table:classification}
\resizebox{1.0\textwidth}{!}{\begin{tabular}{l |c| c }
\hline
\hline
Methods & THUMOS14 & ActivityNet1.2\\
 \hline
 \hline
iDT+FV \cite{wang2013action}& 63.1 &66.5\\
C3D  \cite{tran2015learning}&- &74.1\\
TSN \cite{wang2016temporal} &67.7& 88.8\\

W-TALC \cite{paul2018w} &85.6 &93.2 \\
3C-Net \cite{narayan20193c}&86.9 &92.4 \\
\hline
\hline
Ours  &{\bf87.6}& {\bf93.2} \\
\hline
\hline
\end{tabular}}
\end{minipage}
 \hfill
\makeatletter\def\@captype{figure}\makeatother
\begin{minipage}{0.49\linewidth}
\centering
    \includegraphics[width=5cm,height=4.3cm]{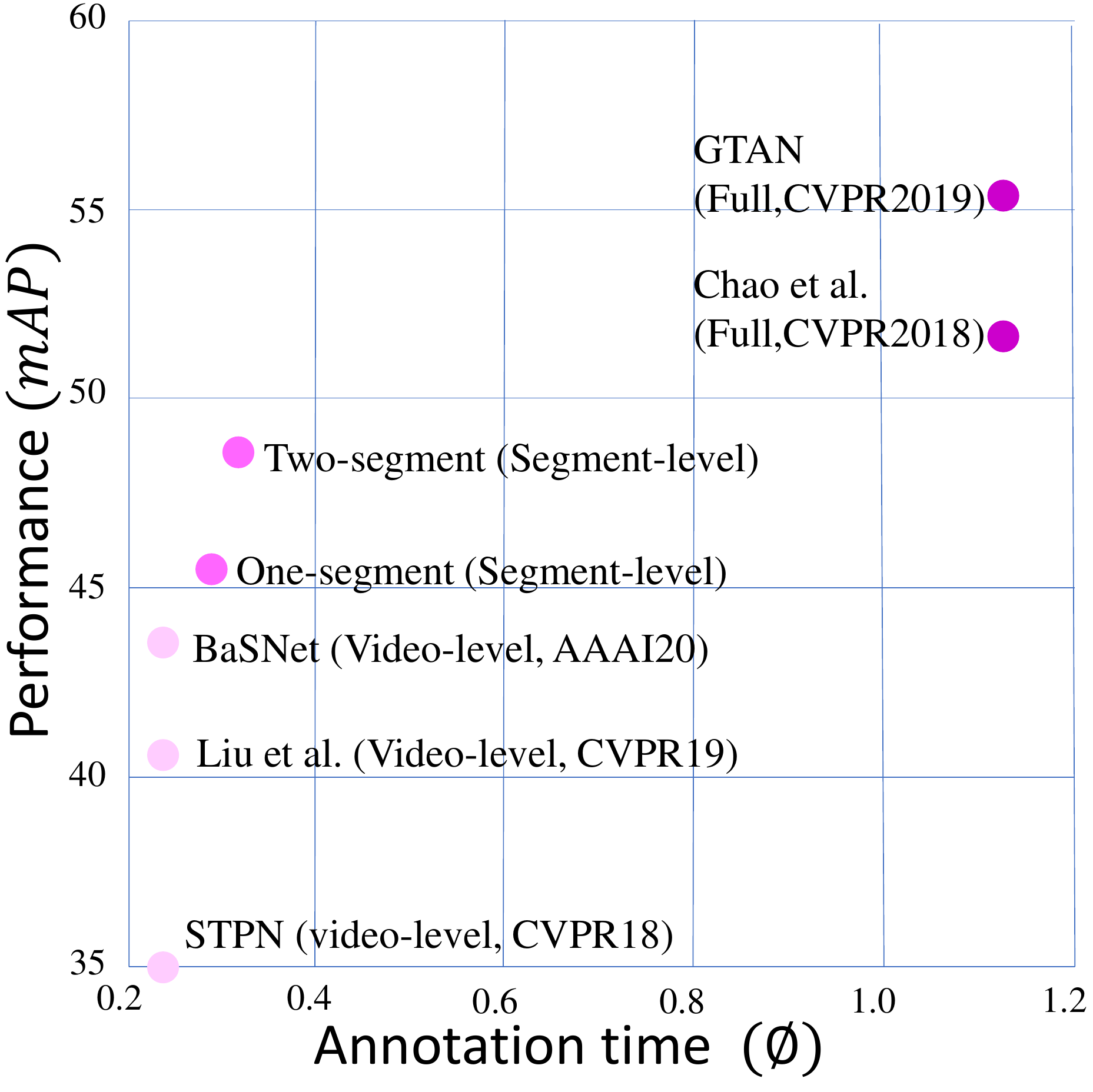}
    \caption{Trade-off of Annotation time and Performance.}
    \label{fig:time_mAP}
\end{minipage}
\end{table}

\begin{figure}[t]
\centering
\includegraphics[width=0.98\columnwidth,height=0.5\textheight]{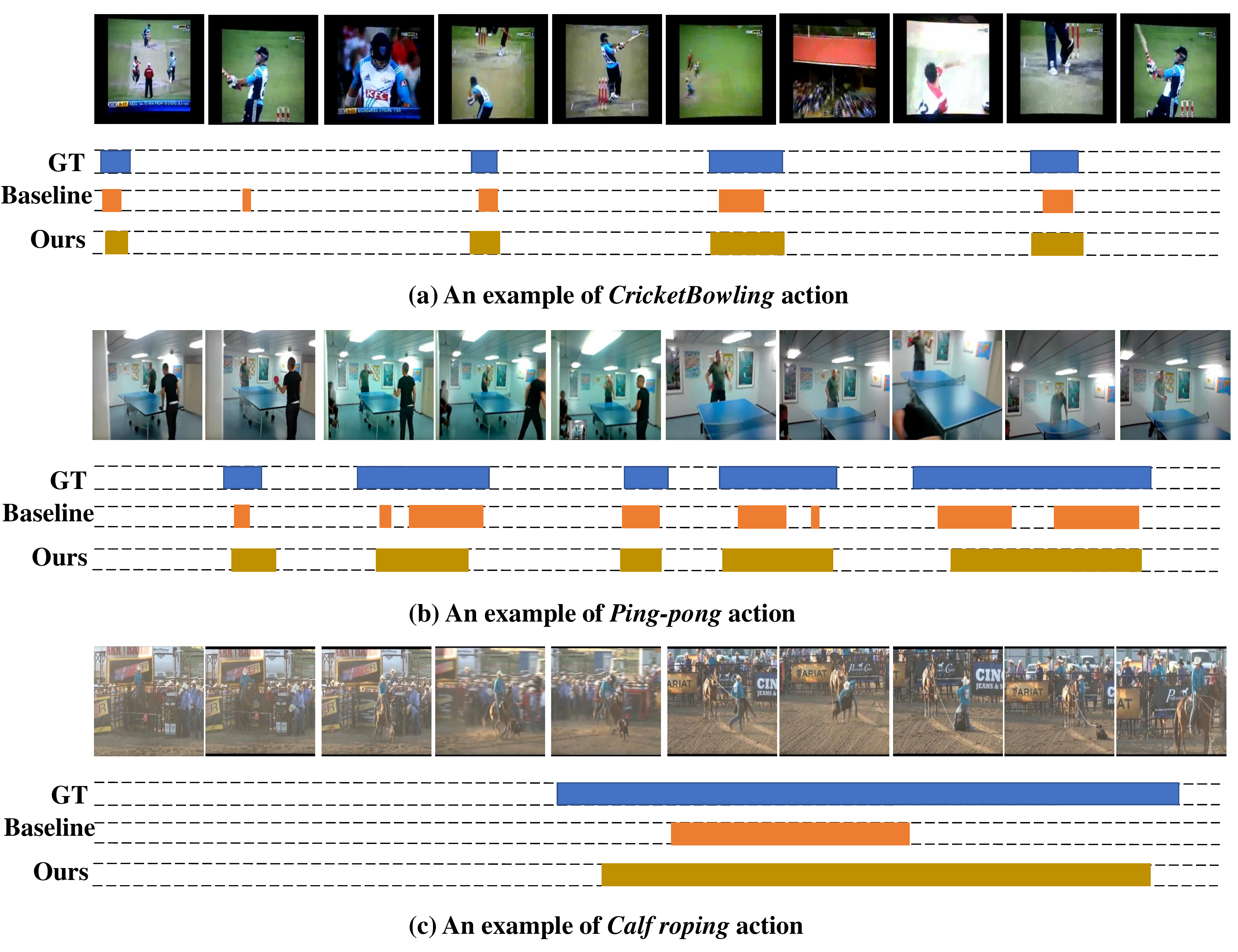}
\caption{Illustration of the temporal predicted regions on the THUMOS14 and ActivityNet datasets. `GT' indicates the ground-truth. `Baseline' means the model only with CL. Our method greatly improves the performance of temporal action localization on all three action videos.}
\label{fig:qualitative}
\end{figure}

\subsection{Qualitative Results}
The qualitative analysis of our approach is shown in Fig. \ref{fig:qualitative}. The top row is the sampling segments of action videos. We choose three actions (`CricketBowling', `Ping-pong' and `Calf roping') from THUMOS14 and ActivityNet to evaluate our method. `GT' denotes the ground-truth segments. The baseline model with only CL just localize the strong discriminative regions. As the duration of time increases, the IoU will decrease. For instances, in Fig. \ref{fig:qualitative} (a), the baseline model can predict segments which have high overlap with GT, for short duration of `CricketBowling'. However, in Fig. \ref{fig:qualitative} (b), a low overlap is obtained for longer duration of `Ping-pong', which is
lower in Fig.~\ref{fig:qualitative}~(c). With our proposed method, more complete and correct action segments will be detected. This indicates that our method can significantly improve weakly supervised temporal action localization.

\section{Conclusion}
\label{sec:conclusion}
In this work, we propose a new segment-level supervision setting for weakly supervised temporal action localization, which costs almost the same annotation time as video-level supervision. Based on the segment-level supervision, we devise a localization module guided by the partial segment loss, the sphere loss and the propagation loss. Compared with video-level supervision, our approach, exploiting the segment labels and propagating them to explicit segments based on the discriminative features, significantly improves the integrity of predicted segments.

	\clearpage
	\bibliographystyle{splncs04}
	\bibliography{reference}

\end{document}